\documentclass[letterpaper, 10 pt, conference]{ieeeconf}  
\IEEEoverridecommandlockouts                              
\usepackage{graphicx,graphics,lipsum}
\usepackage{cite}
\usepackage{comment}
\usepackage{url}
\usepackage{color}			
\usepackage{bm,bbm}
\usepackage{amsmath}
\usepackage{amssymb}
\usepackage{tablefootnote}
\usepackage{float}

\title{\LARGE \bf
	Orientation-Aware Model Predictive Control with Footstep Adaptation for Dynamic Humanoid Walking 
}
\author{Yanran Ding$^1$, Charles Khazoom$^1$, Matthew Chignoli$^1$ and Sangbae Kim$^1$
	\thanks{$^1$The authors are with the Department of Mechanical Engineering at the Massachusetts Institute of Technology, MA - 02139, USA. email: \texttt{\{yanran, ckhaz, chignoli, sangbae\}@mit.edu}}
}

\begin{document}
	\maketitle
	\thispagestyle{empty}
	\pagestyle{empty}
	\begin{abstract}
		This paper proposes a novel orientation-aware model predictive control (MPC) for dynamic humanoid walking that can plan footstep locations online. Instead of a point-mass model, this work uses the augmented single rigid body model (aSRBM) to enable the MPC to leverage orientation dynamics and stepping strategy within a unified optimization framework. With the footstep location as part of the decision variables in the aSRBM, the MPC can reason about stepping within the kinematic constraints. A task-space controller (TSC) tracks the body pose and swing leg references output from the MPC, while exploiting the full-order dynamics of the humanoid. The proposed control framework is suitable for real-time applications since both MPC and TSC are formulated as quadratic programs. Simulation investigations show that the orientation-aware MPC-based framework is more robust against external torque disturbance compared to state-of-the-art controllers using the point mass model, especially when the torso undergoes large angular excursion. The same control framework can also enable the MIT Humanoid to overcome uneven terrains, such as traversing a wave field.
	\end{abstract}
	
	\section{INTRODUCTION}

	The ability to resist unforeseen external disturbance is one of the most critical considerations for bipedal walking. Humanoid robots should leverage multiple recovery strategies, including rapidly planning footstep locations and utilizing torso angular dynamics, to sustain stable walking. However, the periodic switching of foot contact renders the dynamics hybrid \cite{westervelt2003hybrid}, complicating the controller design. Furthermore, foot underactuation and ground contact constraints impose strict limits on the set of feasible reaction wrench the ground can provide. In general, it is challenging to enable a foot-underactuated humanoid robot to simultaneously leverage multiple recovery strategies while fulfilling various constraints in real-time.
	
	\begin{figure}
		\centering
		\resizebox{1\linewidth}{!}{\includegraphics{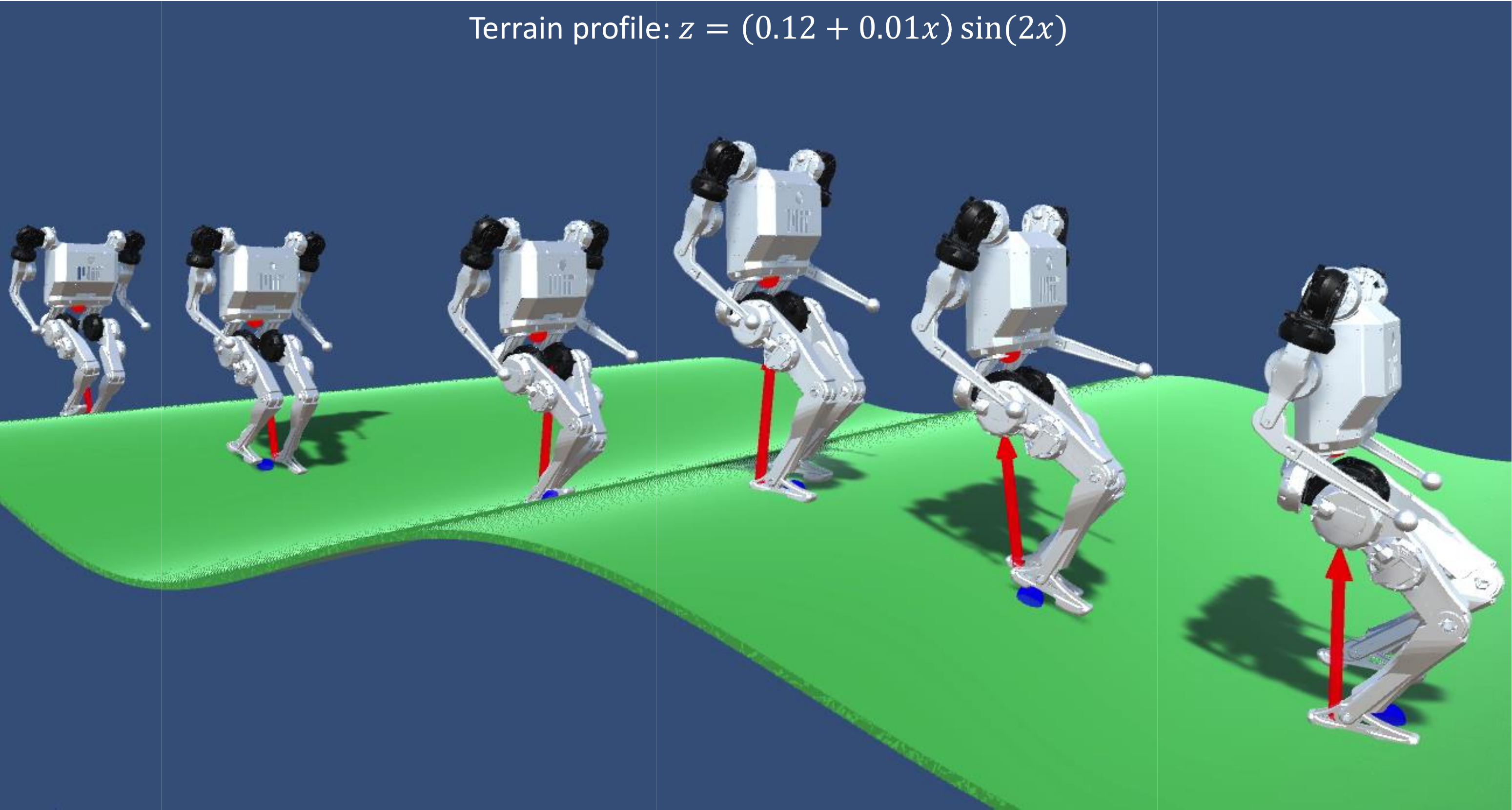}}
		\caption{Snapshots of the MIT Humanoid traversing a wave field. The red arrows show the ground reaction force, and the blue dots on the terrain indicate the desired footstep locations, both of which come from solving the MPC problem.}
		\label{fig:fig1_wavefield}
	\end{figure}
	
    The problem of bipedal locomotion has been widely studied for decades. Hybrid zero dynamics (HZD) approach was proposed to generate stable bipedal walking gait \cite{park2012finite}. Large-scale trajectory optimizations with the full-order robot model are used to search for locally attractive periodic orbits, which are tracked online by feedback controllers. In practice, the walking gait of the robot can become unstable when the deviation from the nominal trajectory exceeds local validity, for example, when the robot is undergoing large disturbances. Recent research in HZD-based control has tried to address this issue of stability under disturbances by finding many periodic orbits and switching between them as the robot is perturbed \cite{hartley2017stabilization}. However, finding the set of periodic orbits that can account for the many different disturbances the robot may experience quickly becomes intractable.

	In contrast to the full-dimensional model of the robot, reduced order models (RoM) are low degree-of-freedom (DoF) representations of the original high DoF systems \cite{full1999templates}. RoMs capture the dominant dynamical effect of the original system but with lower computational cost. This makes them useful for the planning and control of high-DoF legged robots \cite{kuindersma2016optimization}. The Linear Inverted Pendulum Model (LIP)~\cite{kajita1991study} has been widely used for bipedal walking control, where the robot is treated as a point mass moving with constant height. Leveraging the analytical solutions of LIP, center of mass (CoM) trajectories can be synthesized from Zero Moment Point (ZMP)-based control methods~\cite{wieber2016modeling}. In addition to the horizontal CoM motion, the step location is included as part of the decision variables for footstep adaptation in a walking motion generator  using linear Model Predictive Control (MPC) \cite{herdt2010online}. Based on LIP, the concept of divergent component of motion (DCM) \cite{takenaka2009real} is extended in \cite{englsberger2015three} to enable bipedal walking with step adjustment on uneven terrains. Combined step location and timing adaptation strategy is also explored in \cite{griffin2017walking,khadiv2020walking}. For robots without actuated ankle, LIP has also been used to approximate the horizontal CoM dynamics at step-level for designing effective    dynamic walking behaviors \cite{xiong20223,gong2021one}. While the LIP provides closed-form solutions and theoretical guarantees, LIP-based approaches often assume small angular excursion and centroidal momentum that can be regularized by the low-level controller. Few exceptions use the LIP with flywheel model to approximate the inertial effect of the torso and limbs \cite{koolenCapturabilitybasedAnalysisControl2012, aftab2012ankle}. Nevertheless, a set of relatively strict assumptions are usually required, which can be challenging when implementing in practice.
	
	More expressive models such as the centroidal dynamics model has been used in an MPC-based humanoid controller \cite{garcia2021mpc}. Similar to quadruped controllers \cite{DiCarlo_cMPC,ding2019real} that use the Single Rigid Body model (SRBM), the MPC controller in \cite{garcia2021mpc} has the ability to leverage non-trivial torso rotational motion. All of these MPC controllers focus on the modulation of the ground reaction force (GRF), and rely on a footstep strategy that is external to the optimization. This approach has been proven to work well for quadrupedal robots. However, stepping strategy is a crucial component in humanoid control, and studies that incorporate a more expressive model and stepping strategy within a unified framework is not yet well explored. Recent work \cite{romualdi2022online} presents a nonlinear centroidal MPC framework with step adjustment capability on iCub. Nevertheless, the presence of local minima and relatively low control frequency could pose challenges for real-time dynamic locomotion.
	
	The main contribution of this work is an MPC-based humanoid controller that uses the single rigid body model to plan footstep locations online. The augmented SRBM is linearized such that 3D orientation dynamics and footstep location can be handled within a unified optimization framework. The optimization problems in this work are convex quadratic programs (QP) \cite{boyd2004convex}, which can be solved at real-time rates. Our simulation results show that the orientation-aware MPC controller can adapt body rotation and step location simultaneously to withstand higher angular disturbances compared to a hallmark walking pattern generator and a state-of-the-art LIP-based stepping controller.

	The remainder of the paper is organized as follows. Section \ref{sec:background} provides the necessary background; Section \ref{sec:humanoid_control} details the components for the MPC-based control strategy; Section \ref{sec:results} exhibits the results of simulation studies and Section \ref{sec:conclusion} presents the concluding remarks and possible future work.
	
	\section{Background}\label{sec:background}
	\subsection{Hybrid Dynamics Model}
	In this work, the humanoid is assumed to walk with the single support phase for simpler contact schedule. Hence, the humanoid system dynamics is considered a single domain hybrid system. The continuous dynamics concern the rigid body dynamics, and the discrete jump captures the instantaneous change at touchdown. The hybrid system $\Sigma$ is defined as
	\begin{equation}\label{eq:hybrid_dynamics}
		\Sigma:
		\begin{cases}
			& \dot{\bm{x}}=\bm{f}(\bm{x},\bm{u}), t\notin\mathcal{S},\\
			& \bm{x}^+=\Delta(\bm{x}^-), t\in\mathcal{S},
		\end{cases}
	\end{equation}
	where the continuous-time dynamics is $\dot{\bm{x}}=\bm{f}(\bm{x},\bm{u})$; the discrete transition $\bm{x}^+=\Delta(\bm{x}^-)$ maps the state $\bm{x}^-$ to $\bm{x}^+$ on the guard of the hybrid system $\mathcal{S}$. 
	
	\subsection{Humanoid Model}
	The MIT Humanoid \cite{chignoli2021humanoid} is a 24 kg robot with high torque density and high bandwidth control capability. The humanoid model has 24 degrees of freedom (DoF), with 18 actuated DoFs (5 DoFs for each leg and 4 DoFs for each arm) and the floating base coordinate. The robot configuration is described by $\bm{q}=[\bm{q}_b^\top, \bm{q}_a^\top]^\top$, where $\bm{q}_b$ is the unactuated torso pose and $\bm{q}_a$ is the configuration of the actuated joints. The standard dynamic equations of motion are
	\begin{equation}\label{eq:full_dynamics}
		\bm{H}(\bm{q})\ddot{\bm{q}}+\bm{C}(\bm{q},\dot{\bm{q}})=\bm{S}_a^\top\bm{\tau}+\bm{J}_c^{\top}(\bm{q})\bm{u},
	\end{equation}
	where $\bm{H}\in\mathbb{R}^{n\times n}$ is the mass matrix; $\bm{C}(\bm{q},\dot{\bm{q}})$ incorporates the centripetal, Coriolis and gravitational terms; $\bm{u}\in\mathbb{R}^6$ is the ground reaction wrench (GRW) and $\bm{J}_c\in\mathbb{R}^{6\times n}$ is the corresponding contact Jacobian matrix. The matrix $\bm{S}_a=[\bm{0}^{18\times 6}, \bm{1}^{18\times 18}]$ is the selection matrix for the actuated joint torque vector $\bm{\tau}\in\mathbb{R}^{18}$. The humanoid model is constructed using Spatial\_V2~\cite{featherstone2014rigid} with rotor inertia, and will be used for rigid body dynamics simulation and task-space control in Section \ref{sec:task_space_control}.
	
	\subsection{Single Rigid Body Model}
	In this work, the RoM for the MIT humanoid is chosen to be the SRBM, which can model both translational and rotational dynamics. Thanks to the design effort to place the heavy actuators close to the torso~\cite{chignoli2021humanoid}, the dominant dynamical effect can be captured by the SRBM dynamics
	\begin{subequations}\label{eq:SRB_dynamics}
	\begin{align}
	    \ddot{\bm{p}}_c &= \frac{1}{M}\sum_i\bm{F}_i+\bm{a}_g\\
	    \frac{d}{dt}(\bm{I}\bm{\omega}) &= \sum_i \bm{r}_i \times \bm{F}_i+\bm{m}_i\\
	    \dot{\bm{R}} &= \bm{\omega}\times \bm{R}
	\end{align}
	\end{subequations}
	where $M$ is the lumped mass and $\bm{I}$ is the inertial tensor expressed in the world frame; $\bm{a}_g=[0,0,-g]^\top$ is the gravitational acceleration vector. $\bm{p}_c\in\mathbb{R}^3$ is the CoM location; 
	$\bm{R}\in SO(3)$ is the rotation matrix;
	$\bm{\omega}$ is the angular velocity expressed in the body frame.
	$\bm{r}_i$ is the vector from CoM to the $i^{th}$ contact point.
	The contact foot produces the ground reaction wrench (GRW) $\bm{u}=[\bm{F}^\top, \bm{m}^{\top}]^\top$, which consists of force and moment. The SRBM will be modified and used for MPC in the following section.

	\section{Humanoid Control}\label{sec:humanoid_control}
	This section details the components of the proposed orientation-aware MPC-based control framework for dynamic humanoid walking.
	
	\subsection{Augmented SRBM}
	To enable the MPC to reason about foot placement, the SRBM state is augmented with the current step location, and the dynamics become
	\begin{equation}\label{eq:nonlin_dyn}
		\dot{\bm{x}}= \frac{\text{d}}{\text{d}t}\begin{bmatrix}
			\bm{p}_c\\
			\dot{\bm{p}}_c\\
			\bm{\Theta}\\
			\dot{\bm{\Theta}}\\
			\bm{c}
		\end{bmatrix}=\begin{bmatrix}
			\dot{\bm{p}}_c\\
			M^{-1}\bm{F}+\bm{a}_g\\
			\dot{\bm{\Theta}}\\
			\bm{I}^{-1}(\bm{r}\times\bm{F} + \bm{m})\\
			\bm{0}
		\end{bmatrix},
	\end{equation}
	where $\bm{\Theta}\in\mathbb{R}^3$ is the roll-pitch-yaw angle representation of the robot torso, where its time derivative is $\dot{\bm{\Theta}}$; $\bm{r} = \bm{c} - \bm{p}_c$ is the vector from CoM to the current step location, where $\bm{c}\in\mathbb{R}^3$ is the current step location, and it remains stationary during stance.
	
	\subsection{Discrete Linear Dynamics}
	To make the final optimization program convex for the real-time requirement, linearization needs to be performed on the nonlinear dynamics (\ref{eq:nonlin_dyn}). One option is to linearize around the reference trajectory, which is a widely adopted technique. Nevertheless, it is possible that the linearized dynamics lose validity when the robot deviates too much from the reference under persisting disturbances, which is the focus of this work. An alternative is to linearize around the current state and control (operating point). This linearization method provides valid approximation of the dynamics even when the current state is far from the reference. Empirical results suggest that linearization around the operating point can provide more robust behaviors when a quadruped is under large disturbances \cite{ding2021representation}.
	
	The nonlinear dynamics is linearized around the operating point, and the forward Euler integration scheme is employed to obtain the discrete linear dynamics
	\begin{equation}\label{eq:discrete_dynamics_noStep}
		\bm{x}_{k+1} = \bm{A}\bm{x}_{k}+\bm{B}\bm{u}_{k}+\bm{d},
	\end{equation}
	where the subscript $k=0,\cdots,N-1$ indexes the time step within the prediction horizon $N$ of the MPC.
	The matrices $\bm{A},\bm{B},\bm{d}$ are defined as follows
	\begin{equation}
		\begin{aligned}
			& \bm{A} = \bm{1} + T_s\cdot \frac{\partial \bm{f}}{\partial \bm{x}}\bigg\rvert_t\in\mathbb{R}^{n\times n}\\
			& \bm{B} = T_s\cdot \frac{\partial \bm{f}}{\partial \bm{u}}\bigg\rvert_t\in\mathbb{R}^{n\times m}\\
			& \bm{d} = T_s\cdot \bm{f}(\bm{x}_t,\bm{u}_t) - \bm{A}\bm{x}_t - \bm{B}\bm{u}_t\in\mathbb{R}^n, 
		\end{aligned}
	\end{equation}
	where the subscript $(\cdot)|_t$ indicates linearization at the current time. $T_s$ is the prediction time step; $\bm{1}$ is the identity matrix; $\bm{x}_t,\bm{u}_t$ are the state and control vector at the current time, where $\bm{u}_t$ is computed at the previous iteration and applied at the current time.
	
	\subsection{Footstep Placement}\label{sec:step_location_adaptation}
	The main contribution of this work is an MPC framework that can reason about body orientation and foot placement in a unified optimization framework. Equation \eqref{eq:discrete_dynamics_noStep} describes the dynamical effect of ground reaction wrench $\bm{u}$. With the augmented SRBM, the effect of taking a step $\delta\bm{c}$ on the dynamics can be quantified as $\bm{A}_c\delta\bm{c}$, where the stepping vector $\delta\bm{c}$ is part of the optimization variables. The matrix $\bm{A}_c\in\mathbb{R}^{15\times 3}$ is a sub-matrix that equates the last three columns of $\bm{A}$, which characterizes the effect of taking a step $\delta\bm{c}$ on the dynamics.
	
	Since the robot is assumed to walk with only single support phase, the step location $\bm{c}$ can be considered to jump instantaneously at the touchdown event. Since the SRBM assumes no leg mass and inertia, the only discrete transition is $\bm{c}^+=\bm{c}^- + \delta\bm{c}$, where $\delta\bm{c}\in\mathbb{R}^3$ is the stepping vector. The stepping vector $\delta\bm{c}$ appears in the dynamics equation at the time step when a contact switch is scheduled. To encode this discrete transition, a boolean vector $\bm{\eta}\in \{0,1\}^N$ is introduced, where $\eta_k=1$ indicates a scheduled step and $\eta_k=0$ otherwise. The boolean vector $\bm{\eta}$ is determined by the gait schedule, which has fixed timing in this work. The final form of discrete linear dynamics is
	\begin{equation}\label{eq:discrete_dynamics}
		\bm{x}_{k+1} = \bm{A}\bm{x}_{k}+\bm{B}\bm{u}_{k}+\bm{d}+\eta_k\bm{A}_c\delta\bm{c},
	\end{equation}
	which can reason about the effect of both GRW $\bm{u}$ and footstep placement $\delta\bm{c}$ simultaneously.
	
	\subsection{Kinematic Constraints}\label{sec:kinematic_constraints}
	One advantage of having footstep location as part of the optimization variable is that kinematic constraints can be naturally enforced. The step size is bounded to prevent the robot from taking too large a step,
	\begin{equation}\label{eq:step_size}
		|\delta\bm{c}|\leq \delta\bm{c}_{max},
	\end{equation}
	where $|\cdot|$ takes the absolute value element-wise. Furthermore, leg over-extension is prevented by imposing constraints on $\bm{r}$, the vector from the CoM to the step location. At time step $k$, $\bm{r}_k = \bm{c}+\delta\bm{c}-\bm{p}_{c,k}$ is bounded by
	\begin{equation}\label{eq:leg_extension}
		|\bm{r}_k^{x/y}|\leq r_{max}^{x/y} + (1-\eta_k)\cdot M_b, k=1,\cdots, N_{le},
	\end{equation}
	where $r_{max}^{x/y}$ is the maximum displacement projected on the ground plane; $\eta_k$ is the binary indicator as in (\ref{eq:discrete_dynamics}); $M_b$ is a sufficiently large positive number such that constraint (\ref{eq:leg_extension}) is only activated when the robot is scheduled to take a step ($\eta_k=1$). In practice, (\ref{eq:leg_extension}) is imposed for a few steps $N_{le}< N$, where the state prediction from the linearized model is reasonably accurate.
	
	\subsection{Uneven Terrains}\label{sec:uneven_terrain}
	In addition to imposing the kinematic limit, having $\delta\bm{c}$ as part of the decision variables allows the MPC to negotiate sloped terrains. Suppose the terrain nearby the stance foot can be approximated by a plane $\mathcal{P}$, then the following linear equality can be imposed on the new step location
	\begin{equation}\label{eq:terrain_plane}
		\bm{c} + \delta\bm{c}\in \mathcal{P}:=\{\bm{c}\in\mathbb{R}^3|\bm{A}_{\mathcal{P}}\cdot \bm{c}=\bm{b}_{\mathcal{P}}\},
	\end{equation}
	where $\bm{c}$ is the current step location; $\bm{A}_{\mathcal{P}},\bm{b}_{\mathcal{P}}$ parametrizes the plane approximation of the terrain around $\bm{c}$. The terrain approximation plane $\mathcal{P}$ can be updated iteratively, e.g. when the humanoid makes new contacts with the ground, or when vision information has been processed. This flexibility allows the controller to traverse uneven terrains, such as the case shown in Fig. \ref{fig:fig1_wavefield}, where the MIT humanoid traverses a wavefield.
	
	\subsection{Line-Foot Contact}\label{sec:line_foot}
	The MIT Humanoid has under-actuated feet that make line contacts with the ground. For completeness, this section briefly goes through the derivation of the contact wrench cone (CWC) for line contact, which is a special case in~\cite{caron2015stability}. The stance foot can generate a contact wrench except the $roll$ moment in the foot frame $\{F\}$, and the CWC constraints of line-contact have to respect the following,
	\begin{equation}\label{eq:CWC1}
		\begin{aligned}
		    & m_x = 0,\\
			& 0\leq F_z \leq F_z^{max},\\
			& |F_{x/y}|\leq \mu\cdot F_z,\\ 
			& -F_z\cdot l_t \leq  m_y \leq F_z\cdot l_h,
		\end{aligned}
	\end{equation}
	where $\mu$ is the coefficient of friction; $l_h$ and $l_t$ are the lengths from the origin of $\{F\}$ to the heel and toe, respectively; $F_z^{max}$ is the maximum vertical ground reaction force; To derive the constraints on $m_z$, consider the GRW was instead produced by two point forces at the heel $\bm{F}^h$ and toe $\bm{F}^t$. Solving the following equations
	\begin{equation}
		\begin{aligned}
			& F_y = F_y^h + F_y^t, m_y = F_z^{h}\cdot l_h - F_z^{t}\cdot l_t \\
			& F_z = F_z^h + F_z^t, m_z = F_y^{t}\cdot l_t - F_y^{h}\cdot l_h,
		\end{aligned}
	\end{equation}
	and plugging the solutions of $F_y^h,F_y^t,F_z^h,F_z^t$ in the friction constraint $|F_{x,y}^{h/t}|\leq\mu F_z^{h/t}$, one can derive the linear inequality constraints on $m_z$. Collecting all linear inequalities related to line-foot contact to matrices $\bm{A}_{cwc}$ and $\bm{b}_{cwc}$, the CWC set $\mathbb{U}$ can be defined as
	\begin{equation}\label{eq:CWC_constraint}
    	\mathbb{U}:=\{\bm{u}~|~\bm{A}_{cwc} \cdot {^F}\bm{u} \leq \bm{b}_{cwc}\}.
	\end{equation}
	where the left superscript $^F(\cdot)$ indicates variables expressed in the foot frame. 
	
	\subsection{Cost Function}\label{sec:cost_function}
	The MPC cost function $\ell(\bm{x},\bm{u})$ is defined as follows
	\begin{equation}\label{eq:cost_function}
	    \ell(\bm{x},\bm{u}) = ||\bm{x}^d-\bm{x}||_Q^2 + ||\bm{u}^d-\bm{u}||_R^2 + ||\bm{c}^d-\bm{c}-\delta\bm{c}||_{R_c}^2,
	\end{equation}
	where the norm $||\bm{z}||_M^2=\frac{1}{2}\bm{z}^\top\bm{M}\bm{z}$ is the quadratic norm weighted by the matrix $\bm{M}$; the superscript $(\cdot)^d$ indicates the desired values. The reference state $\bm{x}^d$ is a function of the current state and the velocity commands in the $x,y$ and $yaw$ directions; The reference ground reaction force $\bm{u}^d$ points from the stance foot to the CoM, with its vertical force component equal to the total weight.
	
	The proposed MPC framework is flexible in the sense that prior work can be incorporated into the cost function. For instance, the stepping controllers such as capture point \cite{Pratt_2006_capt_pt} and Step-to-Step dynamics-based stepping \cite{xiong20223,gong2021one} can be taken as the reference foot placement $\bm{c}^d$ in (\ref{eq:cost_function}). Since the reference enters the cost term as regulatory terms, the MPC retains the flexibility to optimize the footstep location depending on the current state.

	\subsection{Convex MPC Formulation}
	The cost function and the linear constraints from the previous sections are summarized into the convex MPC, which is transcribed as an optimization problem,
	\begin{equation}
		\begin{aligned}
			\underset{\bm{u}_k,\bm{x}_k,\delta\bm{c}}{\text{min.}}& ~~~~~~ \gamma^{N}\ell_N(\bm{x}_{N}) + \sum_{k=0}^{N-1}\gamma^{k}\ell_k(\bm{x}_{k},\bm{u}_{k})\\
			\text{s.t.}& ~~~~~~ \text{discrete linear dynamics: } (\ref{eq:discrete_dynamics}) \\
			& ~~~~~~ \text{kinematic constraints: } (\ref{eq:step_size}) (\ref{eq:leg_extension})\\
			& ~~~~~~ \text{step location constraint: } (\ref{eq:terrain_plane})\\
			& ~~~~~~ \text{line-contact constraint: }\bm{u}_{k} \in \mathbb{U}, (\ref{eq:CWC_constraint})\\
			& ~~~~~~ \bm{x}_0 = \bm{x}(t),
		\end{aligned}
		\label{eq:convex_MPC}
	\end{equation}
	where the final and stage costs follows the expression described in (\ref{eq:cost_function});  
	the discrete linear dynamics and other constraints are defined in (\ref{eq:discrete_dynamics}-\ref{eq:terrain_plane},\ref{eq:CWC_constraint}). $\bm{x}_0=\bm{x}(t)$ is the initial condition.
	The decay rate $\gamma\in(0,1)$ is a parameter that alleviates the inaccuracy induced by the linearized SRBM dynamics, since more discount is placed on predicted state and control towards the end of the prediction horizon. 
	\par
	The convex MPC formulation (\ref{eq:convex_MPC}) can be transcribed into a QP. Its optimization variable vector is a concatenation of the predicted control and state within the prediction horizon~\cite{wang2009fast}, plus variables that represent step location change 
	\begin{equation}
		[\bm{u}_0^\top,\bm{x}_1^\top,\cdots,\bm{u}_{N-1}^\top,\bm{x}_N^\top,\delta\bm{c}_1^\top,\delta\bm{c}_2^\top]^\top.
	\end{equation}
	
	\begin{figure*}
	    \centering
	    \resizebox{0.95\textwidth}{!}{\includegraphics{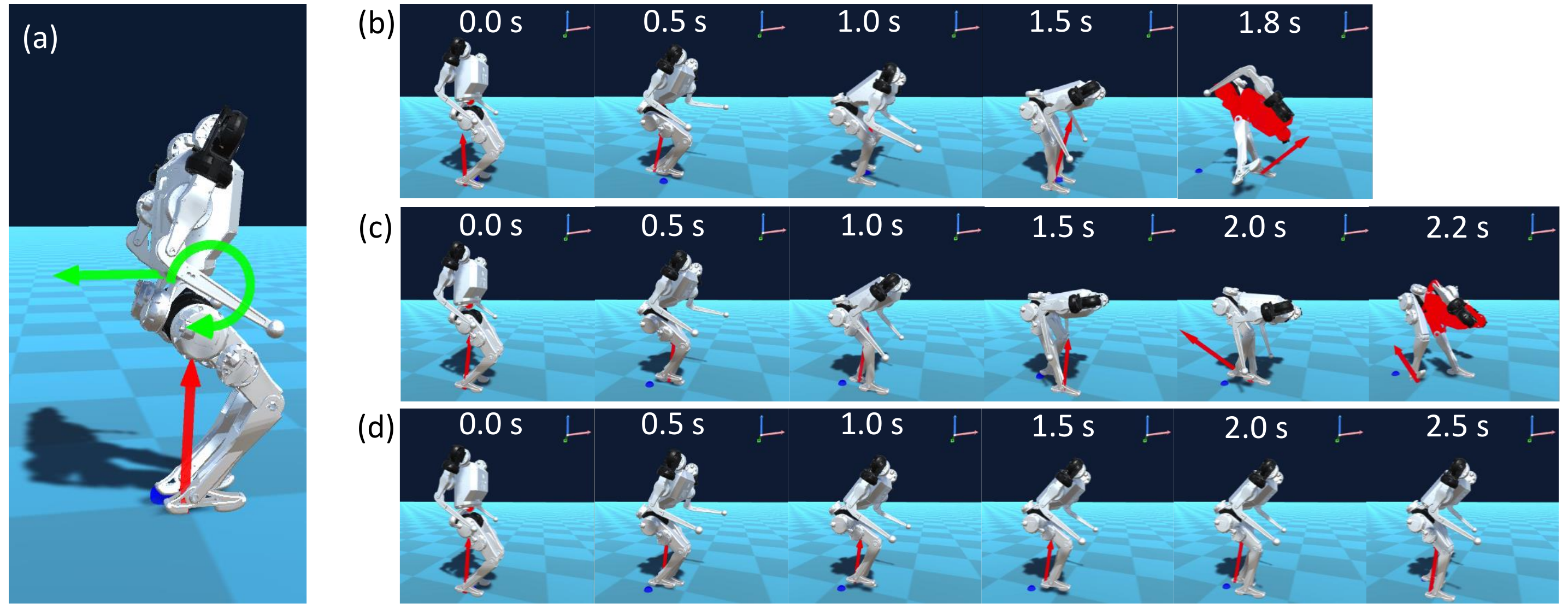}}
	    \caption{Snapshots of the humanoid during the disturbance resistance experiment, where the humanoid is subject to large disturbances that exceed the regulatory capability of the TSC alone. (a) The humanoid is subject to external disturbance wrench (shown as green arrows), which includes both force and torque components ($F_x = -50$ N, $\tau_y=20$ Nm). The red arrow represents the ground reaction force. 
	    (b) The humanoid failed to resist the disturbance wrench using the walking motion generator.
	    (c) The LIP-based controller also could not withstand the torque disturbance.
	    (d) The humanoid survived the disturbance using the orientation-aware MPC controller.}
	    \label{fig:snapshots}
	\end{figure*}
	
	\subsection{Task-Space Control}\label{sec:task_space_control}
	The torso pose, footstep location and contact wrench produced by the MPC are tracked by a Task-Space Controller (TSC). The TSC negotiates various tasks while leveraging the full-body dynamics of the humanoid in equation (\ref{eq:full_dynamics}). A TSC similar to~\cite{wensing2013conic} is implemented here, where joint acceleration $\ddot{\bm{q}}$, joint torque $\bm{\tau}$ and GRW $\bm{u}$ are solved for
	\begin{equation}\label{eq:TSC}
		\begin{aligned}
			\underset{\ddot{\bm{q}},\bm{\tau},\bm{u}}{\text{min.}}& ~~~~~~ ||\bm{A}_t\ddot{\bm{q}}+\dot{\bm{A}}_t\dot{\bm{q}}-\dot{\bm{r}}_t||_{Q_t}^2 + ||\bm{e}_u||_{R_u}^2 + ||\bm{\tau}||_{R_{\tau}}^2\\
			\text{s.t.}& ~~~~~~ \bm{H}\ddot{\bm{q}}+\bm{C}=\bm{S}_a^\top\bm{\tau}+\bm{J}_c^\top \bm{u},\\
			& ~~~~~~ \bm{J}_c \ddot{\bm{q}} + \dot{\bm{J}}_c\dot{\bm{q}} = \bm{0},\\
			& ~~~~~~ \text{line-contact constraint: }\bm{u}\in \bm{\mathbb{U}},\\
			& ~~~~~~ |\bm{\tau}|\leq \bm{\tau}_{max},~|\ddot{\bm{q}}|\leq \ddot{\bm{q}}_{max},
		\end{aligned}
	\end{equation}
	where $\bm{A}_t$ is the task Jacobian matrix; $\dot{\bm{r}}_t$ is the commanded task dynamics; $\bm{e}_u=\bm{u}^d - \bm{u}$ is the deviation from the desired GRW; joint acceleration and torque is regularized through quadratic penalization in the objective, and $\bm{\tau}_{max},\ddot{\bm{q}}_{max}$ bound the joint torques and accelerations.
	\par
	The torso pose and swing foot pose are tracked using a proportional/derivative (PD) control scheme with a feed-forward term in the task space 
	\begin{equation}\label{eq:PD_task_command}
		\dot{\bm{r}}_t = \ddot{\bm{p}}^d + \bm{K}_{P}(\bm{p}^d - \bm{p}) + \bm{K}_{D}(\dot{\bm{p}}^d - \dot{\bm{p}}),
	\end{equation}
	where $\bm{p}$ and $\bm{p}^d$ are the measured and desired task positions, respectively; $\bm{K}_P$ and $\bm{K}_D$ are gain matrices, different for each task. 
	
	The arm joints are regulated to a nominal position except the shoulder pitch joint, whose angle is commanded to be proportional to that of the opposite hip pitch. This heuristic facilitates angular momentum regulation since it encourages arm swing that is in phase with the opposite leg~\cite{3D_slip_wensing_2013}. 
	
	The centroidal angular momentum regulation is another term of the objective in (\ref{eq:TSC}). The commanded rate of change of angular momentum $\dot{\bm{k}}_{G,t}$ is defined as $\dot{\bm{k}}_{G,t} = -\bm{K}_D\bm{k}_G$, since this law dampens excessive angular momentum and promotes balancing \cite{3D_slip_wensing_2013}.
	
	\section{Results}\label{sec:results}
	\subsection{Simulation Setup}
	The simulation experiments are set up as follows: the MPC (\ref{eq:convex_MPC}) is formulated using CasADi~\cite{andersson2019casadi} in MATLAB; the QPs from MPC and TSC are solved using the open-source solver qpSWIFT~\cite{pandala2019qpswift}. An event-based finite state machine is utilized to transition between stance and swing phases. A touchdown event is declared when the swing foot hits the ground, and an impact map is applied to zero the foot velocity. The full-body dynamics of the humanoid are simulated using $ode45$, and the robot motion is visualized in Unity. The weights and other constants of the MPC and TSC can be found in Table \ref{tab:LMPC_weights} and Table \ref{tab:TSC_weights}, respectively.
	
	\subsection{Wave Field}\label{sec:result_wave_field}
\begin{figure}
		\centering
		\resizebox{1\linewidth}{!}{\includegraphics{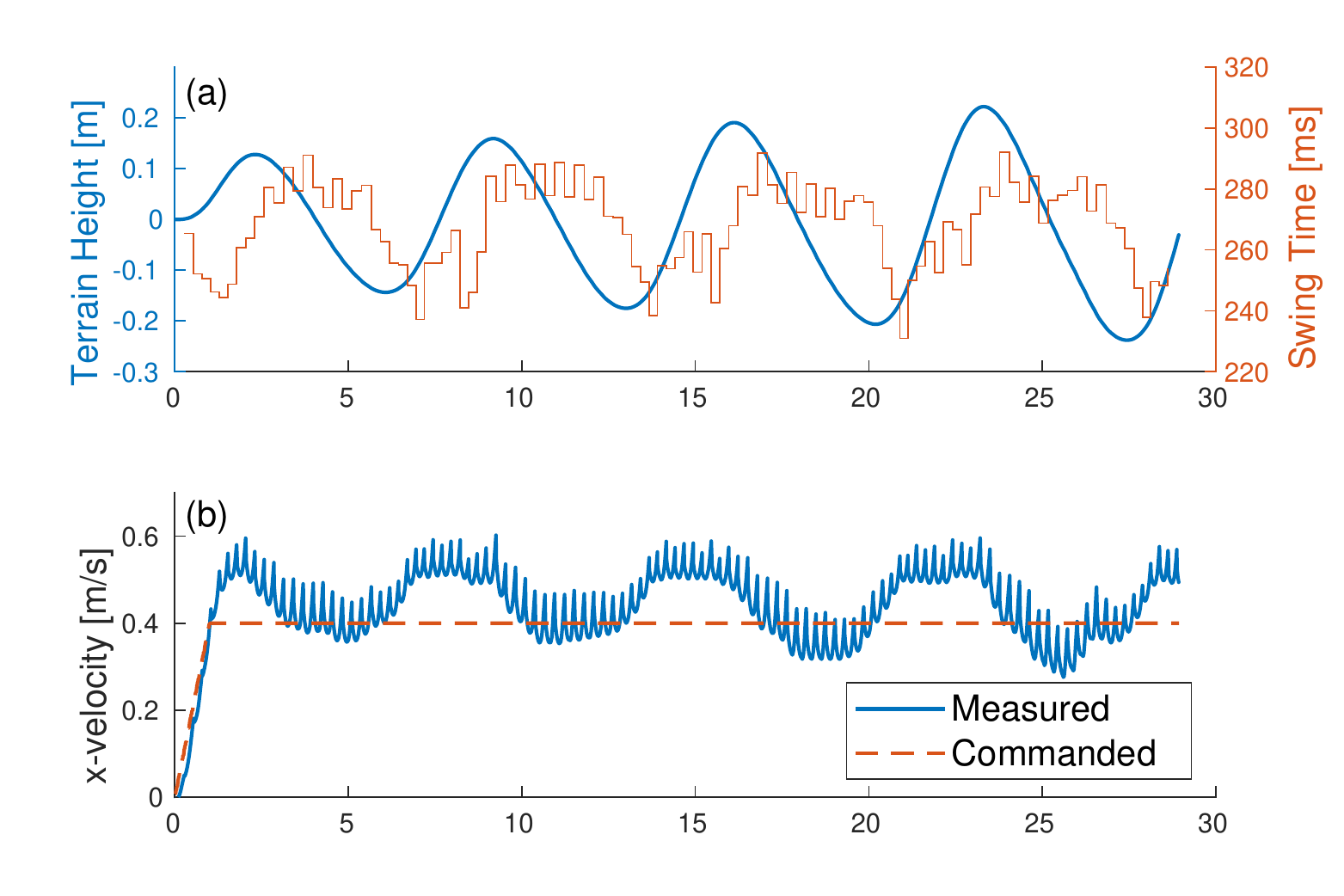}}
		\caption{Simulation result where the MIT Humanoid traverses the wave field shown in Fig. \ref{fig:fig1_wavefield}. (a) The swing time (red) changes with the terrain profile (blue). (b) The commanded (red dashed line) and measured (blue) forward velocity}
		\label{fig:wave_field}
	\end{figure}

	The proposed framework enables the MIT Humanoid to traverse the wave field shown in Fig. \ref{fig:fig1_wavefield}. Since the footstep location is planned on the tangent plane of the current contact, as shown in Section \ref{sec:uneven_terrain}, the swing time deviates from the nominal value depending upon the shape of the terrain. As shown in Fig. \ref{fig:wave_field}(a), where the terrain profile is in blue and swing time in red, the swing foot has an early touchdown where the terrain is convex and a delayed touchdown where the terrain is concave. The maximum slope the robot can traverse is $\pm22^{\circ}$. Fig. \ref{fig:wave_field}(b) shows that the robot can track a commanded velocity of 0.4 m/s.
	
	\subsection{Disturbance Resistance}\label{sec:result_disturbance_recovery}
	A simulation study of disturbance resistance is conducted to highlight the advantages of the orientation-aware MPC. The disturbance resistance experiment is set up such that the humanoid initially stands still on flat ground. A persisting disturbance wrench is applied at the CoM of the humanoid for 2.5 seconds. The disturbance wench consists of a horizontal force $F_x$ and torque along the pitch direction $\tau_y$. Failure is declared if the torso height drops below the threshold (0.3 m). A walking motion generator (WMG) \cite{herdt2010online} that uses robot ankle and a state-of-the-art LIP-based stepping controller \cite{xiong20223} that assumes passive ankle are implemented to the best of our knowledge for the benchmark, where the same TSC with identical gain values is used as the low-level controller for all three methods.
	
\begin{figure}
		\centering
		\resizebox{1\linewidth}{!}{\includegraphics{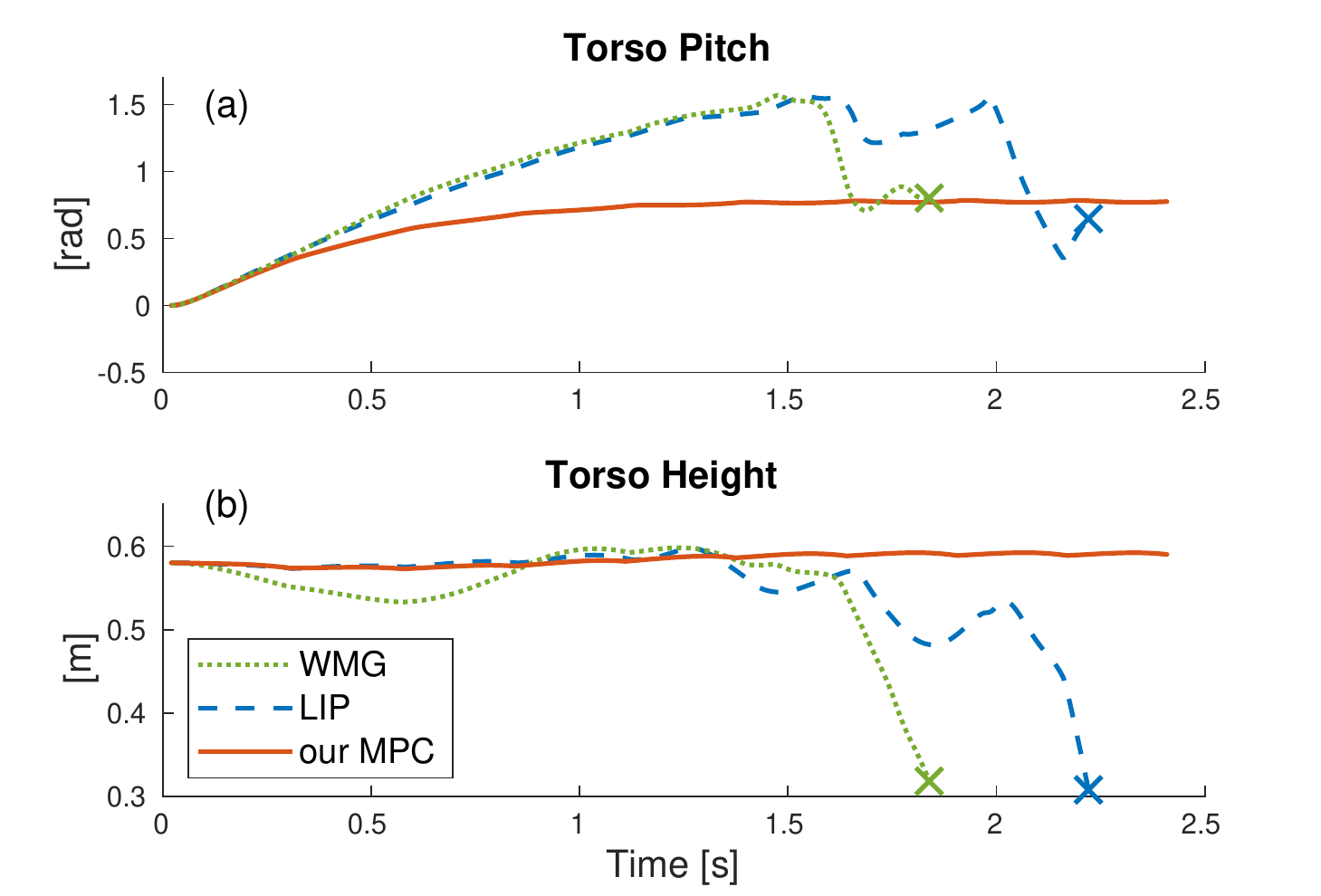}}
		\caption{Simulation data of the disturbance resistance experiment. (a) Torso pitch angle of the humanoid. (b) Torso height of the humanoid with the orientation-aware MPC was maintained around the nominal height, while in the other trials the humanoid fell (indicated by the cross symbols)}
		\label{fig:LMPC_vs_LIP_1try}
	\end{figure}

	Fig. \ref{fig:snapshots} presents the snapshots of the disturbance resistance experiment. Fig. \ref{fig:snapshots}(a) shows that the humanoid is subjected to a pushing force in the negative $x$-direction and a disturbance torque in the positive pitch-direction, as indicated by green arrows. As shown in Fig. \ref{fig:snapshots}(b) and (c), the WMG and LIP controllers can not regulate the high disturbance wrench and the humanoid eventually fell. In contrast, Fig. \ref{fig:snapshots}(d) exhibits that the proposed MPC controller enabled the humanoid to survive the disturbance. It can be deducted that the external wrench has exceeded the regulatory capability of the TSC alone. The ability of the orientation-aware MPC to reason about large angular excursion alleviates the burden of the TSC, and enhance the robustness against large disturbances. Fig. \ref{fig:LMPC_vs_LIP_1try} presents the results of the disturbance resistance experiment, where Fig. \ref{fig:LMPC_vs_LIP_1try}(a) shows that the humanoid torso pitch angle deviated up to 1.3 rad (75$^\circ$) under the WMG and the LIP-based controllers. The CoM height plotted in Fig. \ref{fig:LMPC_vs_LIP_1try}(b) reveals that the proposed MPC controller maintained torso height throughout the experiment whilethe humanoid with the benchmarked controllers fell. That is because the disturbance has exceeded the angular regulation ability of the TSC, unstablizing the WMG and LIP-based controllers. In contrast, the orientation-aware MPC is more robust since it can reason about torso angular dynamics. Furthermore, the capability to plan footsteps within the kinematic constraints avoids leg extension and thus allows the MPC to resist high force and high torque disturbances.
	
\begin{figure}
		\centering
		\resizebox{1\linewidth}{!}{\includegraphics{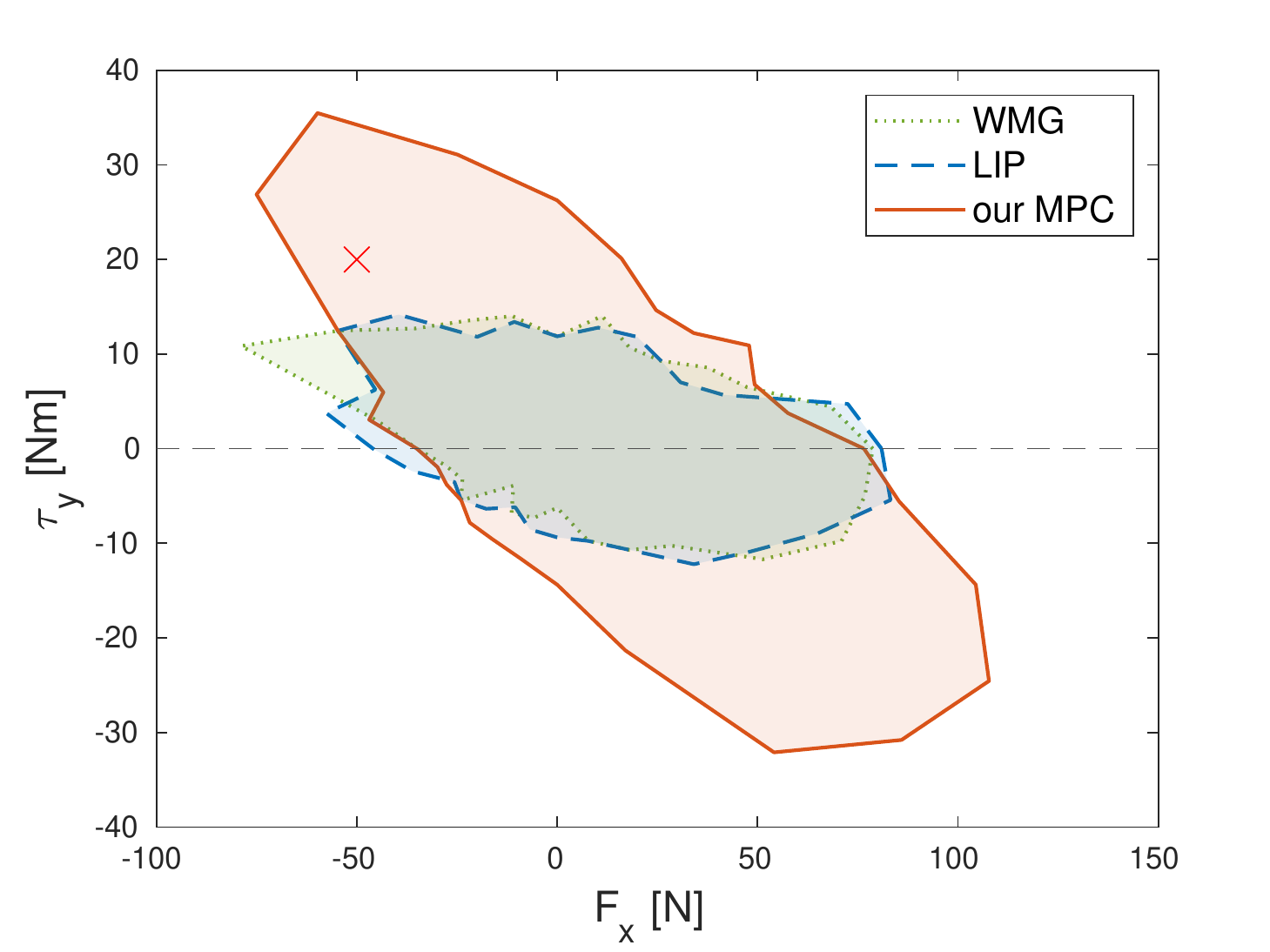}}
		\caption{Comparison of controller performance in terms of resisting persistent disturbance wrench (torque $\tau_y$ and force $F_x$). Three biped controllers are tested, the walking motion generator \cite{herdt2010online} (green), the LIP-based controller \cite{xiong20223} (blue) and the proposed MPC controller (red). The colored regions indicate the set of disturbances that the controllers can recover from. The dashed line indicates the disturbance wrench with only a horizontal force component. The disturbance resistance test case in Fig. \ref{fig:LMPC_vs_LIP_1try} is highlighted by the red cross.}
		\label{fig:LMPC_vs_LIP}
	\end{figure}

	A comprehensive comparison study over a wide range of disturbance wrenches has been conducted. As presented in Fig. \ref{fig:LMPC_vs_LIP}, the walking motion generator and the LIP-based controllers have similar performance in this particular experiment. Both approaches are slightly better than the proposed MPC under horizontal disturbance forces only, as shown by the shaded region near the horizontal dashed line. However, our MPC controller demonstrates a considerable margin of improvement under coupled force-torque disturbances. The orientation-aware MPC performs better under torque disturbances due to its ability to reason about torso rotation and footstep placement simultaneously, while the other methods simplify the robot as a point mass. In summary, the orientation-aware MPC enabled the humanoid to recover from much higher torque disturbances while maintaining comparable performance against horizontal force disturbances.
    
	\begin{table}
		\centering
		\begin{tabular}{c|c|c|c}
			\hline\rule{0pt}{2ex}
			$\bm{p}_c$         & 8e2, 2e3, 3e4 (1)     & $\bm{\Theta}$         & 750, 75, 1250 (0.8)\\
			$\dot{\bm{p}}_c$   & 5e2, 5e3, 5e2 (0.9)   & $\dot{\bm{\Theta}}$   & 8e2, 2e3, 3e4 (0.8)\\
			$\delta\bm{c}$     & 1e5, 1e6, 0           & $\bm{F}$              & 0.01, 0.01, 0.1 (0.7)\\
			$\bm{m}$           & 0.1, 0.1, 0.1 (0.5)   & $N/N_{le}$            & 14 / 3 \\
			$T_s$              & 0.04 s                & $T_{st}$              & 0.38 s \\
			$\delta c_{max}^{x/y}$ & 0.5 m, 0.4 m      & $r_{max}^{x/y}$       & 0.4 m, 0.3 m \\
			\hline
			\multicolumn{4}{l}{\small *values in parentheses are decay rates} \\
		\end{tabular}
		\caption{Weights and other constants for the LMPC}
		\label{tab:LMPC_weights}
	\end{table}
	\begin{table}
		\centering
		\begin{tabular}{l|c|l|c}
			\hline\rule{0pt}{2ex}
			$\bm{K}_P^{t}$          & 100, 100, 100,10, 10, 20      & $\bm{K}_D^{t}$            & 30, 30, 50,50, 50, 50 \\
			$\bm{K}_P^{sw}$         & 350, 350, 560, 70, 70         & $\bm{K}_D^{sw}$           & 10, 10, 17, 7, 7 \\
			$\bm{K}_{P/D}^{q}$      & 1/0.2                         & $\bm{K}_D^{CM}$           & 3 \\
			$Q_{t}$                 & 1                             & $Q_{sw}$                  & 0.7 \\
			$Q_{q}$                 & 0.02                          & $Q_{\ddot{q}}$            & 0.005 \\
			$Q_{CM}$                & 0.3                           & $\bm{R}_{\tau}$           & 1e-8 \\
			$\bm{R}_u$              & 0.01                          & $\ddot{q}_{max}$          & 200 [rad/$s^2$]\\
			\hline
		\end{tabular}
		\caption{Weights and other constants for the Task-space control}
		\label{tab:TSC_weights}
	\end{table}
	
	\section{Conclusion and Future Work}\label{sec:conclusion}
	This paper presents an orientation-aware MPC control framework for dynamic humanoid walking that can simultaneously utilize torso rotation and foot placement for disturbance resistance. The augmented single rigid body model is linearized with respect to the current state, control and step location of the robot, which permits the MPC to reason about the complex coupling between important aspects of humanoid walking in a unified optimization. The outputs from the MPC, including body pose, contact wrench and swing foot pose, are tracked by a task-space controller, which leverages the full-order dynamics of the humanoid. Both the MPC and the task-space control can be transcribed as QPs, which facilitates real-time applications. Simulation experiments are conducted, where the proposed framework is compared with two controllers using the point mass model. It is shown that our method enabled MIT Humanoid to maintain stable walking under substantially higher torque disturbances due to the inclusion of torso angular dynamics and kinematic constraints in the planning. Furthermore, the proposed MPC controller also enables the humanoid to traverse a wave field.
	
	In the future, we plan to make a comprehensive comparison with more existing controllers for push recovery such as \cite{stephens2011push} and \cite{xiong2021robust}. We also plan to improve the linear dynamical model, which currently has the issue of inaccurate prediction towards the end of the horizon. Our future research plan is to refine the the orientation-aware MPC by taking its output as the initial guess to warm start a subsequent nonlinear program. Various techniques such as sequential quadratic program (SQP) and differential dynamic programming (DDP) can be explored to solve the optimization.
	
	\addtolength{\textheight}{-2.5cm}   
	
	\section*{ACKNOWLEDGMENT}
	This work was supported by Naver Labs and the Centers for ME Research and Education at MIT and SUSTech. The authors would like to thank Tyler Matijevich for the insightful discussion.
	
	\bibliographystyle{IEEEtran}	
	\bibliography{IEEEabrv,bibliography}
	
\end{document}